\documentclass[letterpaper]{article} 
\usepackage{aaai2027}  
\usepackage[hyphens]{url}  
\usepackage{graphicx} 
\usepackage{natbib}  
\usepackage{caption} 
\usepackage{algorithm}
\usepackage{algorithmic}
\usepackage{newfloat}
\usepackage{listings}
\DeclareCaptionStyle{ruled}{labelfont=normalfont,labelsep=colon,strut=off} 
\floatstyle{ruled}
\newfloat{listing}{tb}{lst}{}
\floatname{listing}{Listing}
\usepackage{booktabs}
\usepackage{multirow}
\usepackage{amsmath}
\usepackage{amssymb}
\usepackage{pifont}
\usepackage[table]{xcolor}

\title{MT-Web2Code: Benchmarking Coding Agents \\ on Multi-Turn Regional Reconstruction and Localized Modification}

\author{
    Qiming Li\textsuperscript{\rm1,\rm2}, Shujie Hu\textsuperscript{\rm2}, Haohan Liu\textsuperscript{\rm1}, Xiaocheng Feng\textsuperscript{\rm1,3}\corresponding, Songxiang Liu\textsuperscript{\rm2}\corresponding, Guanglu Wan\textsuperscript{\rm2}
}
\affiliations{
    \textsuperscript{\rm 1}Harbin Institute of Technology, \textsuperscript{\rm 2}Meituan, Longcat Team, \textsuperscript{\rm 3}Peng Cheng Laboratory\\
    \texttt{qmli@ir.hit.edu.cn}
}

\begin{document}

\maketitle

\begin{abstract}
Recent advances in Large Vision-Language Models (LVLMs) have demonstrated impressive capabilities in web UI generation. However, existing benchmarks predominantly focus on single-turn full-page generation from scratch, overlooking the iterative workflow of real-world frontend engineering, where developers repeatedly reconstruct missing regions and modify localized elements within existing codebases. To bridge this gap, we introduce \textbf{MT-Web2Code}, the first multimodal coding benchmark for multi-turn \textbf{Macro-Level Regional Reconstruction} and \textbf{Micro-Level Localized Modification}, which contains 102 tasks spanning 16 vertical domains.
To construct deterministic repair trajectories without costly turn-level human annotation, we develop a scalable \textit{Reverse-Corruption Trajectory Engine} that iteratively injects structural and stylistic defects into golden pages. 
We further propose a dual-axis evaluation protocol that measures target-region fidelity and the preservation of unaffected content, where regional reconstruction is assessed by a 5-dimensional VLM-based rubric and localized modification by deterministic pixel-grounded alignment.
Experiments on 13 frontier coding agents reveal that current agents struggle to faithfully reconstruct target regions while preserving unaffected content, lack fine-grained visual-code alignment for localized edits, and suffer from error snowballing over multiple turns.
Beyond benchmarking, our deterministic evaluation metrics provide fine-grained feedback signals that may facilitate future research on training iterative UI coding agents.
Our evaluation code and data will soon be released.
\end{abstract}

\section{Introduction}

Recent advances in Large Vision-Language Models (LVLMs) have enabled increasingly capable multimodal web UI generation \cite{dong2025surveycodegenerationllmbased}. As summarized in Table~\ref{tab:related_work}, existing benchmarks have evaluated these models on a broad range of tasks, including multimodal issue fixing \cite{yang2024swe}, UI-to-code generation \cite{si-etal-2025-design2code,yun2024web2code}, Website Generation \cite{vibe2025, lu2026webgen, dai2026webvr}, full-stack website development \cite{he2026vision2webhierarchicalbenchmarkvisual}, and visual difference detection \cite{zhang2026diffspot}. Despite their diverse task settings, these benchmarks largely share a single-turn formulation: each example is treated as an independent input-output problem, and the model is evaluated on a static final result. Such a formulation can measure whether a model generates a plausible interface, but not whether it can reliably edit an evolving page, preserve previously completed content, or prevent errors from propagating through successive interactions. 
This limitation creates a gap between existing formulations and common iterative frontend editing workflows.
In practice, developers rarely rebuild an entire interface from scratch at every interaction. Instead, they iteratively operate on an existing codebase: reconstructing a missing region, correcting localized structural or stylistic defects, and preserving all content outside the intended scope. 

\begin{figure}[t]
\centering
\includegraphics[width=1\columnwidth]{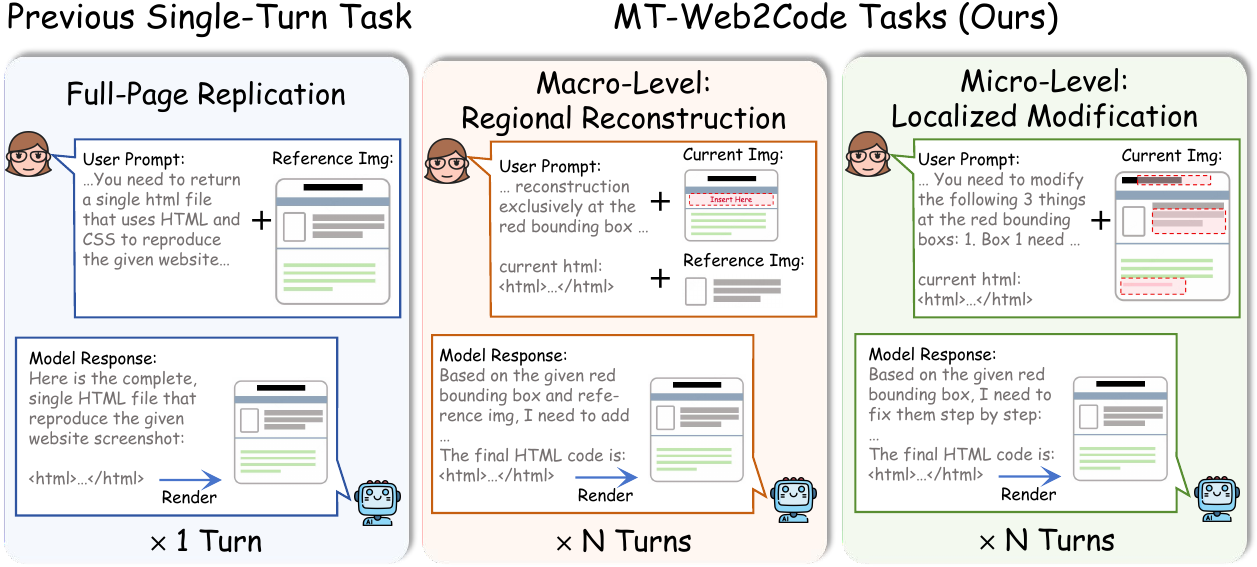}
\caption{Comparison of task formulations. Unlike prior benchmarks that focus on single-turn full-page replication, \textbf{MT-Web2Code} evaluates multi-turn Regional Reconstruction and Localized Modification within existing web pages.}
\label{fig:diff}
\end{figure}

\begin{table*}[t]
\centering
\small
\setlength{\tabcolsep}{5pt}
\begin{tabular}{lccccl}
\toprule
\textbf{Benchmark} & \textbf{Tasks} &\textbf{Domains} & \textbf{Multimodal} & \textbf{Multi-turn} & \textbf{Task Types} \\
\midrule
SWE-Bench Multimodal \cite{yang2024swe} & 617 & - & \ding{51} & \ding{55} &  Issue Fixing\\
Design2Code \cite{si-etal-2025-design2code} &  484  & 8  & \ding{51} & \ding{55} & Single Page UI-to-Code\\
WebGen-Bench \cite{lu2026webgen} &  101  & 20  & \ding{55} & \ding{55} & Website Generation\\
Vision2Web \cite{he2026vision2webhierarchicalbenchmarkvisual}  &   193  & 16 & \ding{51} & \ding{55} & Full-stack Website Development\\
DiffSpot \cite{zhang2026diffspot}    &  4400 & 15 & \ding{51} & \ding{55} & Spot-the-difference on Web Interfaces\\
\midrule
\textbf{MT-Web2Code (Ours)} &102  & 16 &\ding{51} &\ding{51} & Regional and Localized Reconstruction  \\
\bottomrule
\end{tabular}
\caption{Comparison with representative coding and web UI benchmarks. \textbf{MT-Web2Code} evaluates multimodal coding agents on multi-turn \textbf{Macro-Level Regional Reconstruction} and \textbf{Micro-Level Localized Modification}.}
\label{tab:related_work}
\end{table*}

To bridge this gap, we introduce \textbf{MT-Web2Code}, the first multi-turn multimodal coding benchmark dedicated to iterative web UI reconstruction and modification. As illustrated in Figure~\ref{fig:diff}, previous benchmarks ask agents to replicate a complete page in a single turn, whereas \textbf{MT-Web2Code} requires a coding agent to repeatedly modify an evolving page state. The benchmark covers two complementary task granularities. \textbf{Macro-Level Regional Reconstruction} requires the agent to reconstruct an entire missing semantic region from a visual reference and integrate it into the surrounding page structure. \textbf{Micro-Level Localized Modification} requires the agent to repair fine-grained structural and stylistic defects without disrupting unaffected content. Together, these tasks capture the structural reasoning and precise visual-code alignment required in iterative frontend development.

\textbf{MT-Web2Code} contains 102 web pages spanning 16 vertical domains. To construct high-quality multi-turn data without costly turn-level human annotation, we develop a scalable \textit{Reverse-Corruption Trajectory Engine}. Starting from a golden page, the engine injects deterministic structural and stylistic defects and then reverses the corruption sequence into a repair trajectory. This process uniquely defines the target state at every turn, enabling reproducible evaluation while retaining realistic dependencies between successive edits. We further introduce a dual-axis evaluation protocol that measures both target-region fidelity and the preservation of unaffected content, using a five-dimensional VLM-based rubric for regional reconstruction and deterministic pixel-grounded alignment for localized modification.

Extensive experiments on 13 frontier coding agents expose three consistent limitations. First, current agents struggle to faithfully reconstruct missing regions while maintaining consistency outside the target area. Second, they lack the fine-grained visual-code alignment required for precise localized modification. Third, errors occur in early turns are inherited and amplified by subsequent edits, leading to pronounced error snowballing over multi-turn trajectories. These results show that strong performance on static UI generation does not necessarily translate into reliable iterative coding. Beyond enabling dynamic multi-turn evaluation, our deterministic evaluation metrics provide a natural foundation for verifiable reward signals for training iterative UI coding agents.

In summary, our main contributions are:
\begin{itemize}

    \item We introduce \textbf{MT-Web2Code}, the first multi-turn multimodal benchmark tailored for iterative web UI coding, covering regional reconstruction and localized modification across 102 web pages from 16 domains.

    \item We propose a scalable \textit{Reverse-Corruption Trajectory Engine} that constructs deterministic repair trajectories without costly turn-level human annotation, together with a dual-axis protocol that evaluates both target-region fidelity and unaffected-content preservation.
    \item We evaluate 13 frontier coding agents and reveal persistent limitations in regional reconstruction, fine-grained visual-code alignment, and multi-turn consistency.
\end{itemize}

\section{Related Work}

\paragraph{Multi-turn Coding Agents.}
Recent advances in LVLMs have enabled multimodal coding agents with multi-turn reasoning and iterative code execution capabilities. Unlike static code generators, these agents can perceive visual environments, follow natural language instructions, and perform sequential code edits. Recent frontier models, including Claude-4.7-Opus \cite{anthropic2026claudecode}, Kimi-k2.5 \cite{team2026kimi}, and GLM-5V-Turbo \cite{hong2026glm}, demonstrate strong visual-code understanding and instruction-following abilities. However, their multi-turn coding capabilities remain largely underexplored and lack quantitative evaluation.

\paragraph{UI-to-code Generation and Evaluation.}
Recent studies have explored LVLMs for translating UI screenshots into front-end code. Early works leveraged synthetic datasets such as WebSight \cite{laurenccon2024unlocking}, while later efforts, including Web2Code \cite{yun2024web2code} and Flame-React \cite{ge2025advancing}, scaled training with real-world datasets such as WebCode2M \cite{gui2025webcode2m}. Existing benchmarks mainly focus on single-turn whole-page reconstruction. Design2Code \cite{si-etal-2025-design2code} uses global metrics such as Block-Match and CLIP similarity \cite{radford2021learning} to evaluate visual fidelity, while Vision2Web \cite{he2026vision2webhierarchicalbenchmarkvisual} adopts a VLM-as-judge \cite{gu2024survey} approach for overall quality assessment. However, these evaluations cannot measure multi-turn localized editing, which requires both accurate target modification and preservation of unaffected regions. 

\section{MT-Web2Code Bench}
In this section, we first formulate the multi-turn repair task, then detail data preparation and the \textit{Reverse-Corruption} engine for constructing macro- and micro-level trajectories, and finally introduce the corresponding evaluation protocols.
\subsection{Task Formulation}

We formulate \textbf{MT-Web2Code} as a multi-turn trajectory comprising a forward corruption process and an autoregressive reverse repair process. Let $H_0=H^\star$ denote a golden web page, and let $\mathcal{V}(\cdot)$ denote a deterministic rendering function under a fixed coordinate system. The forward process generates a sequence of corrupted states:
\begin{equation}
H_t=C_t(H_{t-1}), \qquad t=1,\ldots,T,
\end{equation}
where $C_t=\{o_{\mathrm{del}}\}$ removes one semantic region in the macro-level track, while $C_t=\{o_1,\ldots,o_K\}$ injects $K$ localized defects in the micro-level track.

The coding agent starts from the fully corrupted state $H'_T=H_T$ and reverses the trajectory autoregressively:
\begin{equation}
H'_{t-1}=R_t(H'_t,\mathcal{I}_t),
\qquad t=T,\ldots,1,
\end{equation}
where $\mathcal{I}_t$ contains the turn-specific instruction and resource assets. Each repair step conditions on the agent's own output from the preceding turn, thereby capturing error propagation across interactions. Since every repair turn reverses a deterministic corruption set $C_t$, its ground-truth target $H_{t-1}$ is uniquely defined, bypassing the evaluation ambiguity inherent in open-ended generation tasks.

\begin{figure}[t]
\centering
\includegraphics[width=0.90\columnwidth]{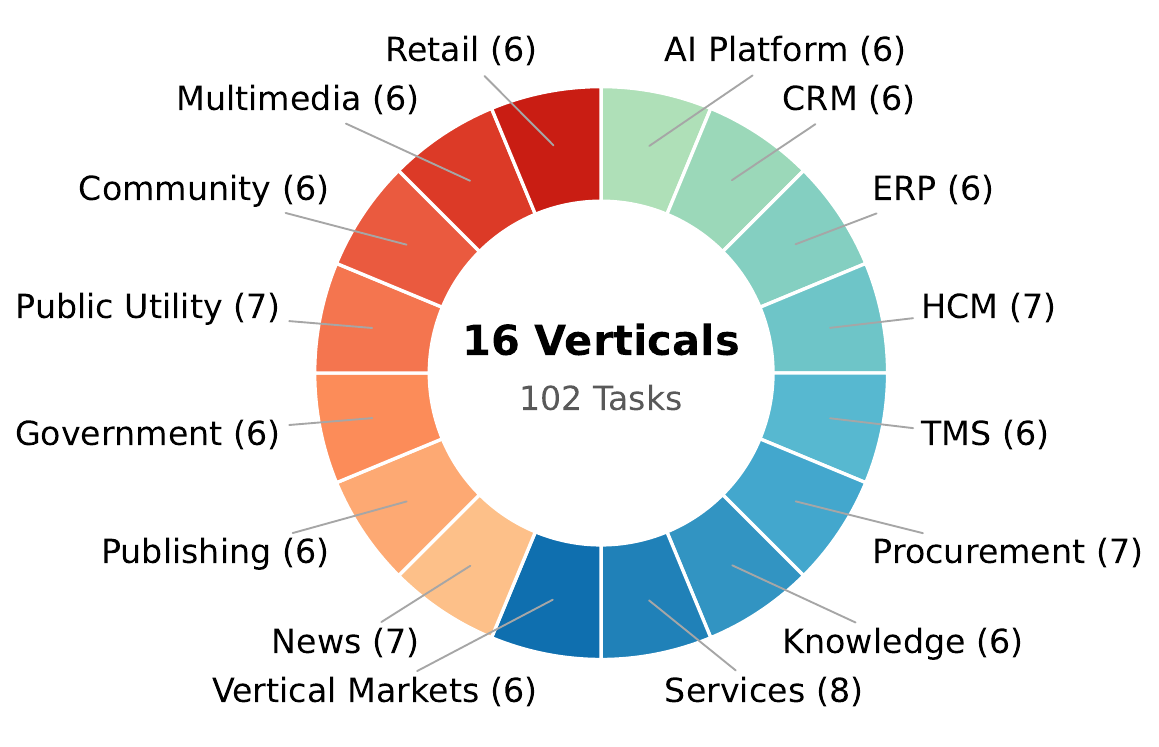}
\caption{The 16 vertical domains of \textbf{MT-Web2Code}, which are sampled to a balanced count.}
\label{fig:domains}
\end{figure}

\subsection{Data Preprocessing}

\paragraph{Web Corpus Collection and Filtering.}
As shown in Figure~\ref{fig:domains}, we initially collect web pages from 16 vertical domains to ensure structural diversity. To construct a high-quality benchmark, these raw pages are rendered in a local environment. We then apply structural and visual similarity filtering to remove redundant or malformed layouts. This screening process results in a final set of 102 high-quality web pages. Each selected page, containing its HTML, CSS, and related resources, serves as a \textit{Golden Page} ($H_0$).

\paragraph{Element Fingerprint Injection.}
A foundational challenge in generating multi-turn trajectories is tracking elements across sequential modifications. Standard DOM tree indices are fragile to structural deletions, and visual coordinates drift during CSS reflows. To address this, prior to the corruption phase, we parse the initial DOM tree and assign a position-agnostic \textit{Element Fingerprint} to each HTML node $e \in H_0$, injected as a custom attribute:
\begin{equation}
\mathrm{ID}(e) = \mathrm{hash}(\mathrm{tag}(e), \mathrm{text}(e), \mathrm{attr}(e))
\end{equation}
This persistent anchor strictly tracks structural boundaries for macro-level regional deletions and enables accurate bounding box localization for micro-level perturbations amid layout reflows. To prevent data leakage, these identifiers are completely removed from the observable states before evaluation by the \textit{Coding Agent}.

\begin{figure*}[!ht]
\centering
\includegraphics[width=\textwidth]{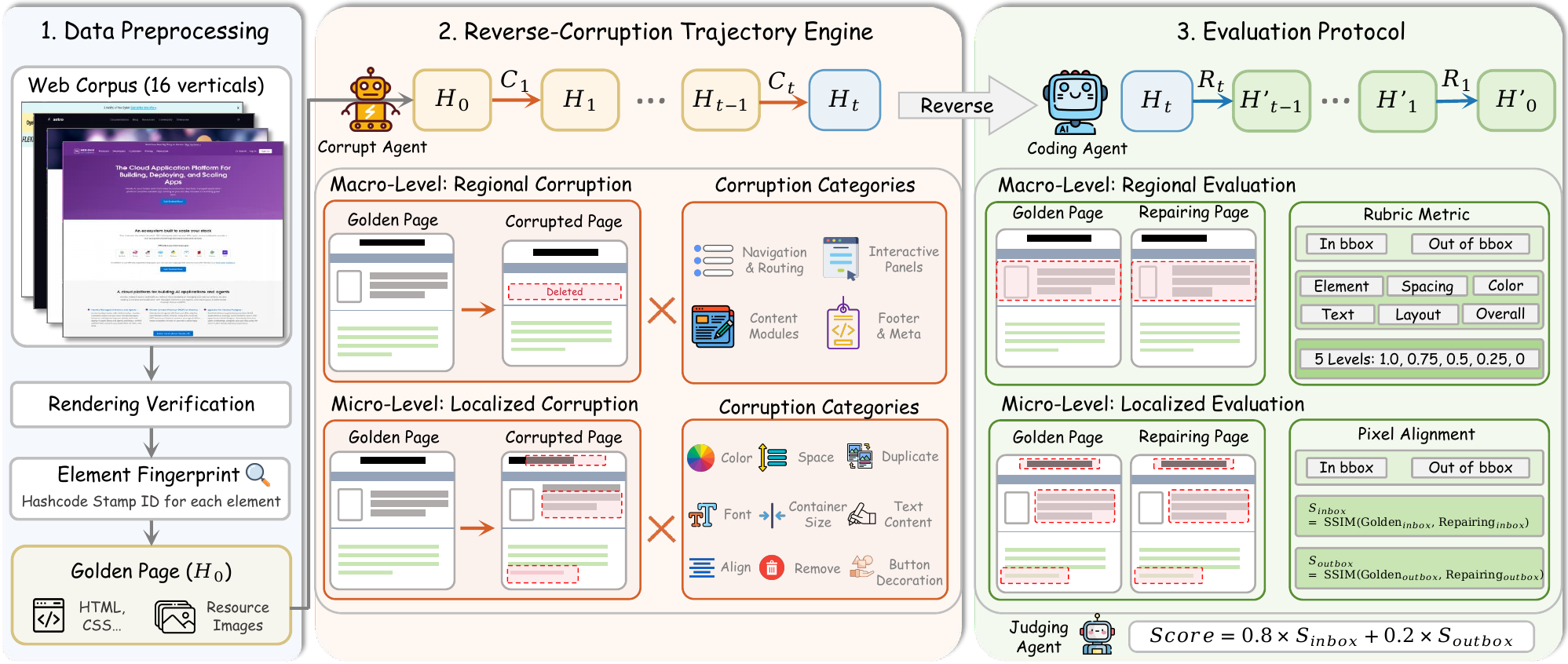}
\caption{Overview of \textbf{MT-Web2Code}. Starting from a golden page, the \textbf{Reverse-Corruption Trajectory Engine} generates two complementary evaluation tracks: \textbf{(I) Macro-Level Regional Reconstruction}, which evaluates regional visual grounding and structural reasoning capability using a rubric-based protocol; and \textbf{(II) Micro-Level Localized Modification}, which evaluates fine-grained visual grounding and visual-code alignment using a pixel alignment protocol.}
\label{fig:overview}
\end{figure*}

\subsection{Macro-Level: Regional Reconstruction}
\label{sec:stage1}
The macro-level task evaluates \textit{Coding Agents} for regional visual grounding and structural reasoning capabilities. At this level, the \textit{Corrupt Agent} executes a structural deletion operation, defined as $C_t = \{o_{\text{del}}\}$, which removes an entire top-level semantic container from the page. Conversely, the \textit{Coding Agent} is tasked with the inverse repair operation, $H'_{t-1} = R_t(H'_t, \mathcal{I}_t)$. This requires the agent to generate HTML and CSS code necessary to restore the missing region and seamlessly integrate it into the required space.

\paragraph{Regional Corrupted Injection.}
To ensure the perturbation is semantically meaningful rather than a random visual crop, the agent targets top-level regions. Let $r \in H_0$ denote a candidate semantic region, structurally represented as a specific node and its enclosed subtree within the Document Object Model (DOM). Guided by the Element Fingerprint, this candidate region $r$ is valid for deletion only if it satisfies:
\begin{equation}
\begin{aligned}
&\mathrm{tag}(r)\in\mathcal{T}_{\mathrm{sem}} \;\wedge\;
w(r) \ge \tau_w \;\wedge\;\\
&\tau_h^{\min} \le h(r) \le \tau_h^{\max} \;\wedge\;
\mathrm{sib}(r) \neq \emptyset,
\end{aligned}
\end{equation}
where $\mathcal{T}_{\mathrm{sem}}$ limits candidates to macro-functional tags (e.g., \texttt{nav}, \texttt{section}) to guarantee semantic importance; $w(r)$ and $h(r)$ strictly bound the rendered dimensions so the deletion is visually substantial yet localized; and $\mathrm{sib}(r) \neq \emptyset$ ensures the presence of surviving siblings to act as anchors.

\paragraph{Corrupted Area Grounding.}
Because web layouts dynamically reflow to close gaps, the spatial coordinates of a deleted region are instantly lost. To prevent positional ambiguity, the \textit{Corrupt Agent} utilizes the hash identifiers to deterministically track the target region's nearest surviving siblings:
\begin{equation}
a^{-} = \mathrm{anchor_{prev}}(r), \quad a^{+} = \mathrm{anchor_{next}}(r)
\end{equation}
Rather than mere theoretical trackers, these surviving elements, $a^{-}$ and $a^{+}$, act as explicit structural markers within the corrupted DOM tree. Their primary purpose is to demarcate the exact insertion bounding box for the \textit{Coding Agent}.

\paragraph{Repair Task Formulation.}
With the exact insertion bounding box established, the regional reconstruction constitutes a multi-turn multimodal dialog task for the \textit{Coding Agent}. At each turn, the observation space comprises:
\begin{itemize}
    \item \textbf{Current Corrupted Visual Context ($\mathcal{V}(H'_t)$):} The rendered full-page screenshot of the current corrupted document, marked with a red bounding box overlay that explicitly localizes the target area requiring reconstruction.
    \item \textbf{Target Visual Reference ($\mathcal{V}(r)$):} The cropped golden rendering of the missing region, serving as the definitive visual ground truth for the target bounding box area.
    \item \textbf{Repair Instruction ($\mathcal{I}_t$):} A strict prompt explicitly commands the agent: \textit{``Do not modify any other content on the page; replicate the target according to the reference image at the position marked by the red bounding box.''} This prompt constrains the generation space, preventing the model from hallucinating global restyling.
    \item \textbf{Resource Assets:} A localized dictionary containing the exact file and their paths of the region's constituent media assets (e.g., local images, SVG icons). 
\end{itemize}
By explicitly coupling the red bounding box with a strict ``no-touch'' textual constraint, the task effectively eliminates arbitrary whole-page hallucinations, forcing the model to focus purely on visual grounding and structural reasoning.

\subsection{Micro-Level Localized Modification}
\label{sec:stage2}
The micro-level track evaluates \textit{Coding Agents} for fine-grained visual grounding and precise visual-code alignment. At this level, the \textit{Corrupt Agent} executes a bundled edit operation $C_t = \{o_1, \dots, o_K\}$ by sampling from 9 atomic operators. These operators are organized into 5 perceptual dimensions $\mathcal{D} = \{\textsc{Layout},\textsc{Elements},\textsc{Text},\textsc{Color},\textsc{Spacing}\}$, summarized in Table \ref{tab:flaws}. This categorization systematically populates the defect space, with operators acting either as a stylistic mutation (\textit{Sty.}) of a single CSS property (leaving DOM topology intact) or a structural edit (\textit{Str.}) to the DOM.

\begin{table}[t]
\centering

\small
\begin{tabular}{lcc}
\toprule
\textbf{Dimension} & \textbf{Atomic Operators} & \textbf{Type} \\
\midrule
Layout    & Alignment (text-align, indentation)     & \textit{Sty.} \\
Elements  & Element Removal, Duplication            & \textit{Str.} \\
Text      & Text-content Edit, Font Size / Weight   & \textit{Sty.\&Str.} \\
Color     & Foreground / Border color               & \textit{Sty.} \\
Spacing   & Margin, Padding, Sizing, Decoration     & \textit{Sty.} \\
\bottomrule
\end{tabular}

\caption{The micro-modification defect taxonomy. Nine atomic operators are organized along five perceptual dimensions. Stylistic(\textit{Sty.}) operators perturb a single rendered property. Structural(\textit{Str.}) operators edit the DOM.}
\label{tab:flaws}
\end{table}

The micro-level track evaluates coding agents for fine-grained visual grounding and precise visual-code alignment. At this level, the \textit{Corrupt Agent} executes a bundled edit operation, defined as $C_t = \{o_1, \dots, o_K\}$, which injects $K$ fine-grained corruptions. Conversely, the \textit{Coding Agent} performs the inverse repair operation, $H'_{t-1} = R_t(H'_t, \mathcal{I}_t)$, requiring targeted HTML/CSS updates to resolve all defects simultaneously while preserving the surrounding layout.

\paragraph{Localized Corrupted Injection.}
To prevent linguistic ambiguity in the repair trajectory, the engine enforces \emph{topological independence} between any two targeted nodes $e_i$ and $e_j$. Given the partial order $\prec$ representing the ancestor-descendant relationship in $T$, we require:
\begin{equation}
\forall\, e_i, e_j \in \{e_1, \dots, e_K\}: 
\\ e_i \prec e_j \implies \mathrm{dim}(o_i) \neq \mathrm{dim}(o_j)
\end{equation}
This constraint prevents cascading repair failures where a structural modification to an ancestor invalidates the visual properties of its descendant, ensuring that each target $e_i$ remains independently repairable.

\paragraph{Visibility Verification.}
An operator $o$ is accepted only if it satisfies a minimum pixel-delta threshold $\tau_{\mathrm{vis}}$:
\begin{equation}
\Delta(H, o) \;=\; \big\| R(H) - R(o(H)) \big\|_{\mathrm{px}} \;\ge\; \tau_{\mathrm{vis}}
\end{equation}
If imperceptible, the operator is resampled until the defect is visually evident.

\paragraph{Corrupted Area Grounding.}
Within a multi-defect step, each edit reflows the page, causing the target's bounding box $b_i$ to drift. We separate \emph{editing} from \emph{coordinate tracking}. Each operator $o_i$ is visibility-checked against the state $H$ just before the edit; its ground-truth bounding box $b_i$ is then re-measured on the fully corrupted page $H'_t$ using the stable Element Fingerprint $\mathrm{id}(e_i)$. This mechanism ensures that even after global layout shifts, the target area for each defect $e_i$ remains deterministic.

\paragraph{Repair Task Formulation.}
With the $K$ defects injected and their drifted coordinates robustly re-measured, the matching repair turn constitutes a parallelized multimodal dialogue task. At each turn, the observation space for the \textit{Coding Agent} comprises:
\begin{itemize}
    \item \textbf{Corrupted Visual Context ($\mathcal{V}(H'_t)$):} The rendered full-page screenshot of the corrupted document, augmented with bounding box overlays. Each bounding box explicitly localizes a specific defective element.
    \item \textbf{Natural Language Instruction ($\mathcal{I}_t$):} A strict directive guiding the agent to resolve the localized issues (e.g., \textit{``Fix the styling and structural defects for the elements marked by the red bounding boxes. Do not modify any other content on the page.''})
    \item \textbf{Resource Assets:} A localized dictionary containing the exact file paths of the page's constituent media assets (e.g., local images, SVG icons).
\end{itemize}
Crucially, unlike the macro-level reconstruction, the agent is not provided with any target visual references (golden crops). This forces the model to rely entirely on its internal prior of web design esthetics, combined with the provided resource assets, to deduce the correct CSS/HTML properties and accurately restore the elements in the highlighted regions.

\subsection{Evaluation Protocol}
\label{sec:eval}

Existing global metrics, including visual similarity measures and VLM-based judgments, provide only coarse assessments of the entire page: they dilute localized errors and fail to distinguish successful repairs from collateral changes to unaffected content. Leveraging deterministic golden references, we therefore introduce a dual-axis protocol that separately evaluates \emph{in-box} fidelity and \emph{out-of-box} preservation for both macro- and micro-level tasks.

\subsubsection{Rubric Metrics for Macro-Region Reconstruction}
Region reconstruction admits many valid HTML realizations of an identical appearance, and is therefore scored perceptually rather than by textual match. The reconstructed region, bracketed on the model's page by the surviving anchors $a^{-}, a^{+}$, is cropped and compared against \textbf{the golden crop} by a VLM judge under a fixed rubric. 

The judge rates fidelity along five equally-weighted dimensions $\mathcal{D} = \{\textsc{Layout}, \textsc{Elements}, \textsc{Text}, \textsc{Color}, \textsc{Spacing}\}$---these dimensions along which Stage-2 defects are organized (Section~\ref{sec:stage2})---each anchored to an explicit five-band scale $\mathcal{S} = \{\textsc{1.0}, \textsc{0.75}, \textsc{0.5}, \textsc{0.25}, \textsc{0.0}\}$, and emits a holistic overall rating $s_{\mathrm{ovr}}$. These six perceptual scores are averaged into an in-box fidelity term:
\begin{equation}
\textbf{$\mathbf{S_{inbox}}$} = \frac{1}{6}\bigg(\sum_{d\in\mathcal{D}} s_d + s_{\mathrm{ovr}}\bigg) 
\end{equation}

Faithful reconstruction must leave the surrounding page intact, as captured by the \emph{out-of-box} term \textbf{$\mathbf{S_{outbox}}$}. With the reconstructed region masked, the VLM judge compares the golden and repaired pages block by block, penalizing collateral reflow or restyling. Structural similarity serves as a reference and fallback when judging fails, while a collapsed or empty region receives an in-box score of zero.

\subsubsection{Pixel Alignment for Micro-Level Modification}
Unlike macro-reconstruction, fine-grained repair possesses a single correct appearance and is accordingly scored deterministically. Each turn is rendered into three images within a shared coordinate system: the golden repaired page $G$, the broken input $B$, and the model output $M$. 

We first define the \emph{repair footprint} $\Omega$---the precise region that legitimately changes---as the pixel-difference mask between $G$ and $B$, which elegantly requires no DOM identifiers:
\begin{equation}
 \Omega = \mathbb{1}\!\big[\,|R(G) - R(B)| > \tau\,\big] 
\end{equation}
Mirroring the macro-level evaluation structure, the deterministic scoring is strictly decoupled into in-box and out-of-box dimensions. The \emph{in-box} repair accuracy is defined as the structural similarity strictly within the footprint:
\begin{equation}
\textbf{$\mathbf{S_{inbox}}$} = \mathrm{SSIM}_{\Omega}(G, M) 
\end{equation}
Conversely, the \emph{out-of-box} structural preservation evaluates the agreement outside the footprint:
\begin{equation}
\textbf{$\mathbf{S_{outbox}}$} = \mathrm{SSIM}_{\bar{\Omega}}(G, M) 
\end{equation}
This out-of-box term is further complemented by a color-sensitive pixel-difference penalty to capture recoloring errors to which structural similarity alone is insensitive.

\paragraph{Unified Score Formulation.}
Regardless of the task granularity—whether relying on perceptual VLM judgments for macro-regions or deterministic pixel alignments for micro-edits—the final score unifies the targeted fix and global preservation using an identical weighting scheme:
\begin{equation}
\textbf{$\mathbf{Score}$} = 0.8\times \textbf{$\mathbf{S_{inbox}}$} + 0.2\times \textbf{$\mathbf{S_{outbox}}$}
\end{equation}

\subsection{Toward Verifiable Feedback Signals}
Our proposed evaluation protocol provides deterministic, reproducible, and dense per-turn rewards grounded in an unambiguous golden state. Therefore, \textbf{MT-Web2Code} extends beyond a static benchmark and provides potential reward signals for future training of iterative UI \textit{Coding Agents}. The same pixel-grounded and rubric metrics used for evaluation can directly serve as training signals for reinforcement learning, eliminating the need for human preference annotations.

\begin{table*}[h]
\renewcommand{\arraystretch}{0.95}
\centering
\small
\setlength{\tabcolsep}{6pt}
\begin{tabular}{@{}cl ccccc c cc c@{}}
\toprule
\textbf{Level} & \textbf{Model} & \textbf{Layout} & \textbf{Element} & \textbf{Text} & \textbf{Color} & \textbf{Spacing} & \textbf{Overall} & \textbf{$\mathbf{S_{inbox}}$} & \textbf{$\mathbf{S_{outbox}}$} & \textbf{Score} \\
\midrule
\multirow{12}{*}{\textbf{Macro-Level}}
 & Gemini-3.5-Flash       & \textbf{65.5} & \textbf{73.5} & \textbf{63.4} & \textbf{69.2} & \underline{59.6} & \textbf{61.5} & \textbf{65.4} & \textbf{65.7} & \textbf{65.5} \\
 & Kimi-K2.6              & \underline{62.9} & 70.2 & \underline{62.1} & \underline{65.9} & \textbf{60.2} & \underline{59.4} & \underline{63.5} & \underline{64.5} & \underline{63.7} \\
 & Claude-4.7-Opus        & 61.9 & 67.2 & 61.5 & 64.4 & \textbf{60.2} & 58.8 & 62.3 & \textbf{65.7} & 63.0 \\
 & Qwen3.5-plus & 62.5 & 69.9 & 61.4 &66.0 &56.9&58.5&62.5&63.8&62.8\\ 
 & Gemini-3.1-Flash-Lite  & 59.0 & 64.9 & 59.1 & 65.1 & 57.2 & 55.7 & 60.2 & 53.5 & 58.8 \\
 & Doubao-Seed-2.0-Pro    & 57.3 & 64.2 & 59.4 & 61.9 & 55.0 & 54.6 & 58.8 & 56.0 & 58.2 \\
 & GPT-5.4 & 58.9 &\underline{70.5} &54.2 &60.6 &51.5&53.2&58.2&53.8&57.3\\
 & GLM-5V-Turbo           & 54.3 & 62.0 & 54.9 & 59.6 & 51.2 & 51.1 & 55.5 & 51.2 & 54.6 \\
 & Gemini-3.1-Pro-Preview & 49.5 & 55.4 & 48.5 & 52.0 & 47.5 & 48.2 & 50.2 & 50.9 & 50.3 \\
 & Doubao-Seed-2.0-Lite   & 48.1 & 54.9 & 44.8 & 54.4 & 44.1 & 43.4 & 48.3 & 35.8 & 45.8 \\
 & Qwen3-VL-Plus          & 35.5 & 44.0 & 40.9 & 43.8 & 33.4 & 34.4 & 38.7 & 31.6 & 37.3 \\
 & GLM-4.6V               & 35.1 & 41.0 & 35.9 & 40.9 & 33.4 & 32.1 & 36.4 & 34.1 & 35.9 \\
 & Qwen3-VL-Flash         & 8.0  & 10.1 & 8.3  & 11.3 & 8.2  & 7.4  & 8.9  & 14.5 & 10.0 \\
\midrule
\multirow{13}{*}{\textbf{Micro-Level}}
 & Doubao-Seed-2.0-Pro    & 59.9 & \underline{83.8} & \textbf{85.9} & \textbf{85.0} & 73.6 & \underline{77.7} & \textbf{79.5} & \textbf{99.2} & \textbf{83.5} \\
 & Qwen3.5-Plus           & \textbf{79.8} & \textbf{84.2} & 77.9 & 81.2 & \textbf{76.6} & \textbf{79.9} & \underline{79.4} & 98.9 & \underline{83.3} \\
 & GPT-5.4 & 66.7 &80.6 &82.7 &\underline{84.5} &72.7 &77.5 &77.9 &\underline{99.1} &82.1 \\
 & Gemini-3.1-Pro-Preview & 66.3 & 82.1 & 77.0 & 83.6 & \underline{74.9} & 76.8 & 77.9 & \underline{99.1} & 82.1 \\
 & GLM-5V-Turbo           & 59.3 & 77.8 & \underline{83.2} & 77.5 & 72.4 & 74.0 & 76.1 & 98.7 & 80.6 \\
 & Gemini-3.5-Flash       & \underline{73.3} & 78.3 & 79.3 & 79.9 & 73.2 & 76.8 & 76.4 & 96.7 & 80.5 \\
 & Gemini-3.1-Flash-Lite  & 46.6 & 69.6 & 78.4 & 81.0 & 74.0 & 69.9 & 73.9 & 98.9 & 78.9 \\
 & Claude-4.7-Opus        & 53.3 & 76.7 & 74.4 & 70.1 & 69.1 & 68.7 & 71.8 & \textbf{99.2} & 77.3 \\
 & Qwen3-VL-Plus          & 66.0 & 66.9 & 73.4 & 78.1 & 68.6 & 70.6 & 70.1 & 98.4 & 75.8 \\
 & Kimi-K2.6{*}              & 73.2 & 71.9 & 76.3 & 70.0 & 66.5 & 71.6 & 70.2 & 95.5 & 75.3 \\
 & Doubao-Seed-2.0-Lite   & 33.3 & 50.4 & 47.9 & 42.8 & 49.1 & 44.7 & 48.2 & 98.0 & 58.1 \\
 & GLM-4.6V               & 33.3 & 41.8 & 34.9 & 23.3 & 30.6 & 32.8 & 33.5 & 94.2 & 45.6 \\
 & Qwen3-VL-Flash         & 6.6  & 23.6 & 27.0 & 25.5 & 19.7 & 20.5 & 22.4 & 88.1 & 35.6 \\
\bottomrule
\end{tabular}
\caption{
Main results on 13 SOTA coding agents. All reported results are averaged over three independent runs. For Macro-Level Reconstruction, scores are obtained via VLM-based rubric evaluation using Kimi-K2.6 as the judge model. For Micro-Level Modification, scores are computed using pixel alignment. Bold and underline indicate the best and second-best results.
}
\label{tab:main}
\end{table*}

\section{Experiments}
\subsection{Evaluation Setup}

\paragraph{Models.}
We evaluate \textbf{MT-Web2Code} with a diverse set of SOTA \textit{Coding Agents}, including 13 widely used LVLM models. The evaluated models cover both proprietary and open-source systems, including Gemini \cite{google2026gemini31pro, gemini3}, Claude \cite{anthropic2026claudecode, anthropic2025claude45}, GPT \cite{openai2026gpt54}, Kimi \cite{kimi2026k25}, GLM \cite{hong2026glm, zai2025glm46v,hong2025glm}, Qwen \cite{bai2025qwen3vl}, and Doubao \cite{bytedance2026seed20} series models. Each agent is evaluated under the same multi-turn interaction protocol without additional task-specific optimization. Specifically, we employ K2.6 as the corrupt and judging agent.

\paragraph{Benchmark Configuration.}
For Macro-Level Regional Reconstruction, we select $53$ web pages, where each task consists of $5$ sequential turns. At each turn, \textit{Coding Agent} is required to reconstruct one missing semantic region while preserving the remaining page content. For Micro-Level Localized Modification, we select $49$ tasks, each containing $5$ turns with $3$ localized modifications per turn. \textit{Coding Agent} must simultaneously resolve the injected defects while avoiding unintended changes outside the target regions.

\paragraph{Task Curation and Evaluation Reliability.}

For Micro-Level task curation, we use Kimi-K2.6 for difficulty filtering and manually check generated instructions, as current VLMs may overlook or misidentify fine-grained defects in full-page renderings \cite{zhang2026diffspot}. Consequently, this curated subset is particularly challenging, with Kimi-K2.6 performing below its general baseline. This perceptual limitation affects only data construction and does not introduce evaluation ambiguity: each modified region is compared with its golden crop within a deterministic corruption footprint, enabling precise regional scoring.

\subsection{Main Results}

Table~\ref{tab:main} presents the overall performance of 13 \textit{Coding Agents} on \textbf{MT-Web2Code}. We observe three key findings:

\paragraph{Coding Agents Struggle with Regional Reconstruction.}
On the Macro-Level tasks, current coding agents demonstrate promising but still limited multi-turn UI editing capabilities. Gemini-3.5-Flash achieves the best overall performance with a score of 65.5, followed by Kimi-K2.6 with 63.7. Despite their strong visual understanding capabilities, existing agents still struggle with faithful regional reconstruction. In particular, lightweight models in the Gemini series outperform some larger models, such as Claude-4.7-Opus and Gemini-3.1-Pro-Preview. This is because larger models tend to introduce unnecessary global layout changes during local reconstruction, even when explicitly instructed to preserve unaffected regions. This reveals two key limitations: current \textit{Coding Agents} not only fail to faithfully reconstruct the target region at the specified location, but also fail to consistently preserve unaffected regions during localized editing. 


\paragraph{Agents Preserve Surrounding Content but Fail on Precise Localized Modification.}
On the Micro-Level modification task, agents face a different challenge from regional reconstruction. While most advanced models achieve high out-of-box preservation scores (often above 0.95), their in-box repair accuracy varies substantially. This indicates current agents are generally effective at avoiding unintended modifications but remain limited in fine-grained visual grounding and precise visual-code alignment.

\paragraph{Regional Reconstruction and Localized Modification Require Different Capabilities.}
The complementary performance patterns across the two tasks further validate the necessity of our dual-track evaluation design. Macro-Level reconstruction primarily evaluates structural reasoning and contextual layout understanding, whereas Micro-Level modification focuses on localized visual grounding and precise code editing. For example, Gemini-3.5-Flash achieves the best performance on Macro-Level reconstruction (65.5), while Doubao-Seed-2.0-Pro obtains the highest Micro-Level score (83.5), suggesting that strong page-level reconstruction ability does not necessarily imply superior fine-grained editing capability. These two settings therefore capture distinct aspects of multi-turn UI coding ability that cannot be fully characterized by a single evaluation protocol.

\begin{table*}[!ht]
\centering
\renewcommand{\arraystretch}{1.0}
\small
\setlength{\tabcolsep}{7pt}

\begin{tabular}{ll ccccc c cc c}
\toprule
\textbf{Model} & \textbf{Setting} & \textbf{Layout} & \textbf{Element} & \textbf{Text} & \textbf{Color} & \textbf{Spacing} & \textbf{Overall} & \textbf{$\mathbf{S_{inbox}}$} & \textbf{$\mathbf{S_{outbox}}$} & \textbf{Score} \\
\midrule
                 & w/o Caption & 65.5 & 73.5 & 63.4 & 69.2 & 59.6 & 61.5 & 65.4 & 65.7 & 65.5 \\
Gemini-3.5-Flash & w/ Caption  & 72.3 & 82.6 & 69.9 & 74.6 & 64.6 & 68.4 & 72.1 & 70.9 & 71.8 \\
                 & Diff.       & +6.8 & +9.1 & +6.5 & +5.4 & +5.0 & +6.9 & +6.7 & +5.2 & +6.3 \\
\midrule
                 & w/o Caption & 62.9 & 70.2 & 62.1 & 65.9 & 60.2 & 59.4 & 63.5 & 64.5 & 63.7 \\
Kimi-K2.6        & w/ Caption  & 57.2 & 66.8 & 59.2 & 62.3 & 55.2 & 55.2 & 59.3 & 58.7 & 59.2 \\
                 & Diff.       & -5.7 & -3.4 & -2.9 & -3.6 & -5.0 & -4.2 & -4.2 & -5.8 & -4.5 \\
\midrule
                 & w/o Caption & 61.9 & 67.2 & 61.5 & 64.4 & 60.2 & 58.8 & 62.3 & 65.7 & 63.0 \\
Claude-4.7-Opus  & w/ Caption  & 57.2 & 64.1 & 55.9 & 60.1 & 53.9 & 54.3 & 57.6 & 62.8 & 58.6 \\
                 & Diff.       & -4.7 & -3.1 & -5.6 & -4.3 & -6.3 & -4.5 & -4.7 & -2.9 & -4.4 \\
\bottomrule
\end{tabular}
\caption{Ablation study on caption of regional visual reference. The Diff. rows indicate the performance change (\%) when providing captions of regional visual reference compared to the non-caption baseline. Results are averaged over three runs.}
\label{tab:ablation_caption}
\end{table*}

\begin{table}[t]
\centering

\small
\setlength{\tabcolsep}{6pt}
\renewcommand{\arraystretch}{1.0}
\begin{tabular}{cccc}
\toprule
Judge Model & $\Delta$mean & Full-mark & Corr. \\
\midrule
Kimi-K2.6 (ref.)       & ---     & $\sim$20\% & 1.0 \\
Gemini-3.5-Flash       & $+12.1$ & 50\%       & 0.91 \\
Gemini-3.1-Pro-Preview & $+6.7$  & $\sim$20\% & 0.86 \\
\bottomrule
\end{tabular}
\caption{Judge cross-validation on K2.6 reconstructions.}
\label{tab:judge}
\end{table}

\begin{figure}[!ht]
\centering
\includegraphics[width=1\columnwidth]{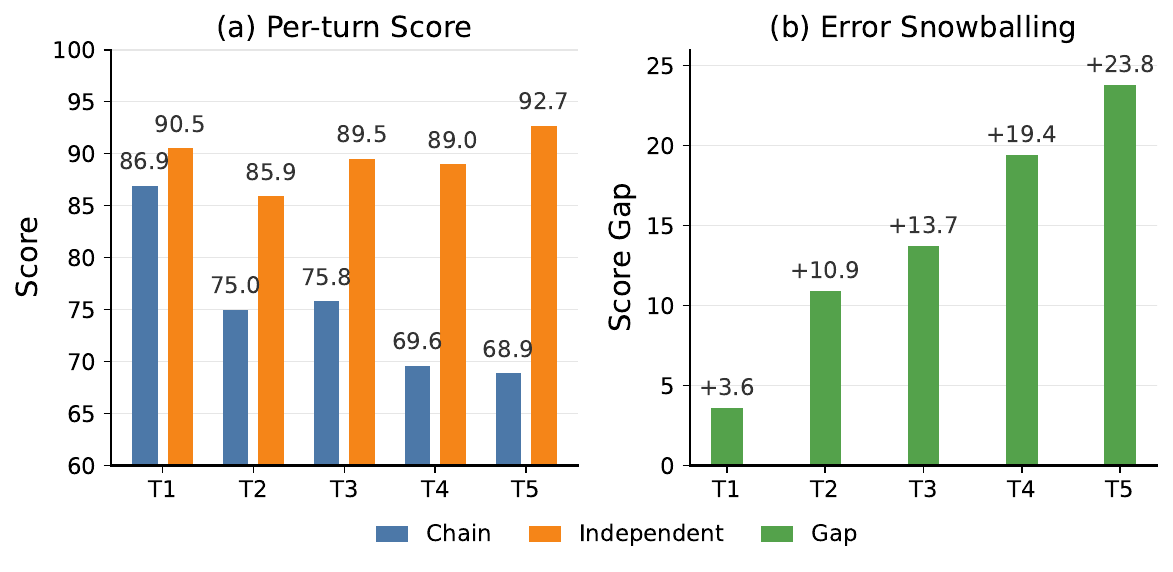}
\caption{Per-turn performance comparison of Kimi-K2.6 between chain-based (default) and independent editing, along with the accumulated error gap across editing rounds.}
\label{fig:snowballing}
\end{figure}

\section{Fine-Grained Analysis}

\subsection{Caption May Hurt Visual Reconstruction}

We investigate whether additional captions of regional visual reference can improve reconstruction by comparing caption and non-caption settings. As shown in Table~\ref{tab:ablation_caption}, textual descriptions provide inconsistent benefits: Gemini-3.5-Flash improves by $+6.3$ points, whereas Kimi-K2.6 and Claude-4.7-Opus degrade by $-4.5$ and $-4.4$, respectively. This suggests that textual guidance may introduce a competing semantic prior that conflicts with visual evidence. Since captions often abstract away fine-grained spatial relationships, component hierarchy, and styling details, models may over-rely on linguistic descriptions and deviate from the actual visual structure. Therefore, faithful UI reconstruction requires grounding generation on visual-code alignment rather than simply incorporating additional textual information.

\subsection{Strong Coders Aren't Always Reliable Judges}
We find that coding performance does not necessarily imply judging reliability. Although Gemini-3.5-Flash achieves the strongest reconstruction results, it tends to assign overly optimistic scores. In contrast, Kimi-K2.6 provides stricter, more discriminative evaluations that better align with human perception, while maintaining competitive coding ability. 
This suggests that judge selection requires fine-grained perception capability more than coding, and a high-performing coding agent is not necessarily an effective judge.

\subsection{Multi-Turn Editing Risks Error Snowballing}

To investigate whether errors accumulate during multi-turn editing, we compare the standard chain-based setting with an independent setting, where each turn receives the corresponding golden previous state instead of the model-generated output from the previous turn. Tasks at each turns have comparable task difficulty, allowing us to isolate the impact of error propagation. As shown in Fig.~\ref{fig:snowballing}, although models achieve similar performance when each turn starts from a correct state, their quality degrades substantially in the chain-based setting as editing progresses. This indicates that early-stage imperfections are continuously inherited and amplified in later rounds, causing an error snowballing effect. Therefore, multi-turn editing introduces additional challenges beyond individual edit capability, where maintaining consistency with previous modifications becomes critical for reliable long-horizon UI generation.

\section{Conclusion}
We introduced \textbf{MT-Web2Code}, the first multi-turn multimodal benchmark for iterative web UI coding, covering regional reconstruction and localized modification. Its \textit{Reverse-Corruption Trajectory Engine} constructs deterministic repair trajectories, while the dual-axis protocol evaluates both repair fidelity and unaffected-content preservation. Experiments on 13 frontier coding agents reveal persistent limitations in regional reconstruction, fine-grained visual-code alignment, and cross-turn consistency under realistic long-horizon multi-turn interaction settings. Beyond evaluation, \textbf{MT-Web2Code} also provides potential verifiable reward signals for future UI coding agents training.

\bibliography{aaai2027}

\end{document}